\documentclass[11pt]{article}

\usepackage[]{acl}

\usepackage{times}
\usepackage{graphicx}
\usepackage{latexsym}
\usepackage[T1]{fontenc}
\usepackage[utf8]{inputenc}
\usepackage{microtype}
\usepackage{multirow}

\newcommand{\name}{\texttt{DictaRashi}}
\newcommand{\repourl}{https://github.com/Dicta-Israel-Center-for-Text-Analysis/dicta_ocr}

\title{\name: A State-of-the-Art OCR Model For Rashi Script}

\author{Ometz Shmidman\textsuperscript{1,*} \quad Yonah Taurog\textsuperscript{1,*} \quad Aryeh Sanders\textsuperscript{1} \quad Shlomo Bluth\textsuperscript{1} \\
\textbf{Moshe Koppel\textsuperscript{1} \quad Joshua Guedalia\textsuperscript{1} \quad Avi Shmidman\textsuperscript{1,2}} \\
\textsuperscript{1}DICTA / Jerusalem, Israel \quad
\textsuperscript{2}Bar Ilan University / Ramat Gan, Israel \\
\texttt{\small \{ometzshmidman,jonah.taurog1,aryehsanders,lomibluth,moishk,joshuaguedalia\}@gmail.com} \\
\texttt{\small avi.shmidman@biu.ac.il} \\
{\small \textsuperscript{*}Equal contribution}}

\begin{document}
\maketitle
\begin{abstract}
We release {\name}, a Tesseract OCR model for scanned Hebrew books printed in the so-called "Rashi" font. We trained {\name} on over one million manually corrected lines of text from over 100 different books of Rabbinic and post-Rabbinic literature. On a held-out set of 1,120 manually transcribed lines from scanned Rashi-script pages, {\name} reaches a character error rate (CER) of 0.86\% and a word error rate (WER) of 3.38\%, setting a new state-of-the-art for publicly-available Tesseract models. We provide a full comparison of our results to those of the 31 other publicly available Hebrew Tesseract models; the best of those models achieves a 2.98\% CER and 11.86\% WER. We further demonstrate that {\name} outperforms the current frontier LLMs on this OCR task. We release the test set of Rashi-script documents so that it may be used as a benchmark for future work in this area. We are pleased to release the model and test set to the community under a CC BY NC 4.0 license.\footnote{Although the official license is limited to non-commercial use, users may contact us to obtain a free license for commercial use in the context of a small business initiative.}
\end{abstract}

\section{Introduction}

A large percentage of published Rabbinic and post-Rabbinic books are set in the so-called "Rashi" script, a semi-cursive Hebrew typeface. Millions of such pages are available as bitmap scans, but full-text search, digital editions and downstream NLP all depend on OCR, and Rashi script is poorly served by general-purpose engines: the official Tesseract Hebrew models read most Rashi letters as their square-script look-alikes and their character-level error rate is generally no lower than 30\%  (Section~\ref{sec:results}).

Several community models fine-tuned for Rashi script exist (Section~\ref{sec:baselines}), but they have not previously been compared on a common, manually verified test set. On this backdrop, this technical report makes three key contributions:
\begin{itemize}
\item We release {\name}, a new Tesseract model trained on over one million manually corrected lines from over 100 Rabbinic and post-Rabbinic books, runnable with a stock Tesseract installation.
\item We create and release a new benchmark for Rashi-script OCR evaluation, consisting of 1,120 lines scanned text from a wide assortment of Rabbinic and post-Rabbinic books, along with ground truth prepared by domain experts.
\item We present a comprehensive comparison and evaluation of 32 open Hebrew Tesseract models as well as 8 leading LLMs on the aforementioned test set of 1,120 hand-transcribed Rashi-script lines.
\end{itemize}

\section{Model and Training}
\label{sec:training}

\subsection{Starting point}
We performed our train as a continuous train on top of the standard \texttt{heb} model from \texttt{tessdata\_best} (version string \texttt{4.00.00alpha:\allowbreak heb:\allowbreak synth20170629}), keeping its network unchanged:
\begin{center}\small\texttt{[1,36,0,1 Ct3,3,16 Mp3,3 Lfys48 Lfx96 Lrx96 Lfx192 O1c1]}\end{center}

\subsection{Training data}
Table~\ref{tab:data} summarizes the data. Pages reserved as test pages for each book are excluded. 

\begin{table}[t]
\centering
\small
\begin{tabular}{|l|r|}
\hline
Books & 118 \\ \hline
Pages & 14,195 \\ \hline
Text lines & 1,343,425 \\ \hline
Words & 12,724,738 \\ \hline
Characters (non-space) & 52,517,573 \\ \hline
Distinct word forms & 389,194 \\ \hline
Pages per book (median) & 96 \\ \hline
Words per page (median) & 821 \\ \hline
\end{tabular}
\caption{Training data.}
\label{tab:data}
\end{table}

\subsection{Procedure}
Our training used the \texttt{tesstrain} scripts\footnote{\url{https://github.com/tesseract-ocr/tesstrain}} with Tesseract \texttt{5.2.0-51-ga8735}. 
Each training sample is one text line image paired with its transcription, rendered as a single line (page segmentation mode 7). \texttt{Tesstrain} was run in fine-tuning mode (\texttt{START\_MODEL=heb}) with \texttt{LANG\_TYPE=RTL}, which implies Unicode normalization mode 3 and builds the language model with \texttt{--pass\_through\_recoder --lang\_is\_rtl}. The pipeline:
\begin{enumerate}
\item concatenates all transcriptions and builds the character set;
\item converts every line transcription into a box file and every image/box pair into an \texttt{.lstmf} sample;
\item shuffles the samples and splits them into training and evaluation lists;
\item runs \texttt{lstmtraining} from the \texttt{heb} checkpoint.
\end{enumerate}
Training ran for 500,000 iterations with the default \texttt{tesstrain} learning rate. The released model is the final checkpoint in floating-point form.

\section{Test Corpus}
\label{sec:testcorpus}

\subsection{Test pages}
The test set consists of 14 Rashi-script pages from typical out-of-copyright Rabbinic and post-Rabbinic books, scanned anew from physical volumes for this evaluation using a Fujitsu ScanSnap SV600 overhead scanner. We have ensured that there is no duplication between the test pages and the training set. Altogether, the pages of the test set contain 1,120 lines, 11,294 words, and 51,749 characters.\footnote{Running headers and page numbers are not scored.}

\subsection{Ground truth}
Domain experts prepared transcriptions for each of these lines using a web annotation tool.

\subsection{Line Images and Bounding Boxes}
The pages are divided into 1,120 lines via Tesseract's layout segmentation routine, using the "auto" setting (PSM=3), which provides bounding boxes for each of the lines. These bounding boxes are considered canonical for the purposes of the test set; when evaluating each OCR system, we request transcription of the lines one at a time, always with these same bounding boxes. 

\subsection{Test Corpus Release Package}
The test set release includes the page images, the bounding boxes and the ground-truth transcriptions.

\section{Evaluation Setup}
\label{sec:setup}

\subsection{Metrics}
We measure CER (character error rate) and WER (word error rate) for each evaluated model across the test corpus. Before scoring, both sides are Unicode NFC-normalized; vowel points, cantillation marks and invisible bidirectional or zero-width characters are removed; every character other than Hebrew letters, digits and spaces is dropped; and whitespace runs are collapsed.

\subsection{Baselines}
\label{sec:baselines}
We compare our model to 31 publicly-available models Hebrew Tesseract models, including the following:
\begin{itemize}
\item the official \texttt{tessdata}, \texttt{tessdata\_best} and \texttt{tessdata\_fast} models, both \texttt{heb} and the \texttt{script/Hebrew} model;\footnote{\url{https://tesseract-ocr.github.io/tessdoc/Data-Files}}
\item three historical official releases;\footnote{\url{https://tesseract-ocr.github.io/tessdoc/tess3/Data-Files.html}}
\item HeOCR R6, R7 and R8b;\footnote{\url{https://huggingface.co/HeOCR}}
\item Rivok, all published releases;\footnote{\url{https://github.com/RivoksLab/tesseract-heb-rivok}}
\item Pninim Rashi;\footnote{\url{https://gitlab.com/pninim.org/tessdata_heb_rashi}}
\item Boro11204;\footnote{\url{https://github.com/boro11204/hebrew-ocr-offline}}
\item EastEriq Rashi.\footnote{\url{https://github.com/tesseract-ocr/langdata/issues/82}}
\end{itemize}

\subsection{LLMs}
We also compare our model's performance to that of the leading frontier commercial LLMs. We compare to eight such LLMs, with reasoning effort set to medium, and with the following prompt: <transcribe exactly as printed, without adding vowel points>.\footnote{The calls to the LLMs were performed via OpenRouter, during November 2026} Each model received each line image in a separate request, just as the Tesseract models did.\footnote{Running the material through the LLMs one line at a time serves an additional purpose: it prevents the model from simply generating its memorized version of the text, due to the limited context of the line. Thus, the test serves to evaluate the LLM's ability to decipher the pixels, rather than its ability to retrieve a pre-known text.} Empty responses count as empty output.

\section{Results}
\label{sec:results}

\begin{table*}[t]
\centering
\small
\begin{tabular}{|l|l|c|c|}
\hline
\textbf{Provider} & \textbf{Model} & \textbf{CER} & \textbf{WER} \\ \hline
Dicta & {\name} (this release) & \textbf{0.86} & \textbf{3.38} \\ \hline
\multirow{5}{*}{Boro11204} & heb\_rashi7 & 2.98 & 11.86 \\ \cline{2-4}
 & heb\_rashi6 & 3.12 & 12.33 \\ \cline{2-4}
 & heb\_rashi5 & 3.52 & 14.32 \\ \cline{2-4}
 & heb\_rashi4 & 5.84 & 23.19 \\ \cline{2-4}
 & heb4 & 22.82 & 66.04 \\ \hline
\multirow{10}{*}{Rivok} & v2.0 heb\_rivok & 5.84 & 20.84 \\ \cline{2-4}
 & v2.0 heb\_rivok (best) & 5.87 & 20.96 \\ \cline{2-4}
 & v2.0 heb\_rivok\_lite (best) & 6.95 & 24.36 \\ \cline{2-4}
 & v2.0 heb\_rivok\_lite & 7.08 & 24.59 \\ \cline{2-4}
 & v1.0.1 heb\_rivok (best) & 20.96 & 60.44 \\ \cline{2-4}
 & v1.0 heb\_rivok (best) & 20.96 & 60.44 \\ \cline{2-4}
 & v1.0.1 heb\_rivok & 21.10 & 60.55 \\ \cline{2-4}
 & v1.0 heb\_rivok & 21.10 & 60.55 \\ \cline{2-4}
 & v1.0.1 heb\_rivok\_lite (best) & 27.05 & 71.80 \\ \cline{2-4}
 & v1.0.1 heb\_rivok\_lite & 27.27 & 72.15 \\ \hline
\multirow{2}{*}{Pninim Rashi} & heb\_rashi & 13.57 & 42.84 \\ \cline{2-4}
 & heb\_rashi\_fast & 13.65 & 43.02 \\ \hline
\multirow{2}{*}{EastEriq Rashi} & heb-ras & 29.02 & 80.64 \\ \cline{2-4}
 & heb-rashi-stam4 & 58.12 & 129.21 \\ \hline
\multirow{6}{*}{Official Tesseract} & tessdata\_best heb & 33.73 & 82.81 \\ \cline{2-4}
 & tessdata heb & 34.28 & 83.81 \\ \cline{2-4}
 & tessdata\_fast heb & 34.93 & 83.95 \\ \cline{2-4}
 & tessdata script/Hebrew & 37.38 & 86.09 \\ \cline{2-4}
 & tessdata\_best script/Hebrew & 37.52 & 86.09 \\ \cline{2-4}
 & tessdata\_fast script/Hebrew & 37.70 & 86.03 \\ \hline
\multirow{3}{*}{HeOCR} & r6 & 35.44 & 81.86 \\ \cline{2-4}
 & r7 & 35.73 & 82.50 \\ \cline{2-4}
 & r8b & 36.67 & 82.85 \\ \hline
\multirow{3}{*}{Tesseract historical} & 4.00 heb & 38.99 & 86.90 \\ \cline{2-4}
 & 3.04.00 heb & 45.23 & 88.49 \\ \cline{2-4}
 & initial (2014) heb & 52.03 & 98.05 \\ \hline
\end{tabular}
\caption{CER and WER (\%, lower is better) of all 32 open Tesseract Hebrew models on the 1,120 lines of our test set. Providers are ordered by their best CER; ``(best)'' marks Rivok's full-precision \texttt{tessdata\_best}-based variants. Note that in one case WER goes above 100\%, because extra words count as errors, and thus, a model which outputs more words than exist in the ground truth may end up with a WER of over 100\%.}
\label{tab:tesseract}
\end{table*}

\begin{table}[t]
\centering
\small
\begin{tabular}{|l|l|c|c|}
\hline
\textbf{Provider} & \textbf{Model} & \textbf{CER} & \textbf{WER} \\ \hline
Anthropic & Claude Fable 5.2 & 2.90 & 8.46 \\ \hline
Google & Gemini 4 Argon & 3.85 & 9.48 \\ \hline
Anthropic & Claude Opus 5.5 & 4.42 & 11.81 \\ \hline
Alibaba & Qwen4 Max & 7.50 & 21.96 \\ \hline
Google & Gemini 4 Hydrogen & 10.20 & 24.92 \\ \hline
OpenAI & GPT-6.2 Astra & 13.70 & 32.42 \\ \hline
xAI & Grok 4.7 & 14.60 & 37.30 \\ \hline
Anthropic & Claude Sonnet 5.5 & 18.30 & 38.74 \\ \hline
\hline
Dicta & \name{} & \textbf{0.86} & \textbf{3.38} \\ \hline
\end{tabular}
\caption{CER and WER (\%, lower is better) of eight frontier LLMs on the same 1,120 line images, with {\name} for reference.}
\label{tab:llm}
\end{table}

\paragraph{Tesseract models.} Table~\ref{tab:tesseract} provides comparative results for all models. {\name} reaches 0.86\% CER and 3.38\% WER, against 2.98\% / 11.86\% for the next best model (Boro11204 \texttt{heb\_rashi7}): about three and a half times fewer errors in both characters and words. The official Tesseract models, HeOCR and the historical models stay above 30\% CER: they were trained on square script and read Rashi letters as their square-script look-alikes.

\paragraph{LLMs} Table~\ref{tab:llm} shows a wide spread across LLMs. The best performing LLM is Claude Fable 5.2 which reaches 2.90\% CER and 8.46\% WER, still substantially below the performance of {\name}. Gemini 4 Argon (3.85\% / 9.48\%) and Claude Opus 5.5 (4.42\% / 11.81\%) follow; the remaining five models range from 7.50\% to 18.30\% CER.

\begin{table}[t]
\centering
\small
\begin{tabular}{|l|l|r|r|r|}
\hline
\textbf{A} & \textbf{B} & \textbf{A$\to$B} & \textbf{B$\to$A} & \textbf{Total} \\ \hline
vav & yod & 30 & 4 & 34 \\ \hline
het & tav & 15 & 12 & 27 \\ \hline
het & pe & 15 & 0 & 15 \\ \hline
kaf & nun & 1 & 13 & 14 \\ \hline
gimel & nun & 0 & 13 & 13 \\ \hline
bet & nun & 7 & 6 & 13 \\ \hline
final mem & samekh & 3 & 9 & 12 \\ \hline
bet & gimel & 9 & 0 & 9 \\ \hline
he & final mem & 0 & 7 & 7 \\ \hline
alef & tav & 1 & 5 & 6 \\ \hline
\end{tabular}
\caption{The ten most frequent letter confusions of {\name} on the 1,120 test lines. A$\to$B: ground-truth letter A read as B.}
\label{tab:confusions}
\end{table}

\paragraph{Error analysis} The most frequent errors in the {\name} results are between visually close Rashi letters (Table~\ref{tab:confusions}): \textit{vav} read as \textit{yod}, \textit{het} and \textit{tav} in both directions, and \textit{het} read as \textit{pe}. The ten pairs in the table account for nearly one third of the errors.

\section{Usage}
The model is a standard \texttt{.traineddata} file and needs no code beyond Tesseract~5:    
\begin{verbatim}
tesseract page.png out \
  --tessdata-dir ./models \
  -l DictaRashi --psm 3
\end{verbatim}
The model, the test set and usage examples are available on Github.\footnote{\url{\repourl}}

\section{Interactive Website}
We provide an interactive website where users can upload scanned Rashi-script documents and receive OCR results from our {\name} model.\footnote{\url{https://rashiocr.dicta.org.il/}}

\section{Future Work}
We are currently pursing additional avenues to achieve even higher accuracy on Rashi-script books using transformer-based models such as TrOCR.

\section{Conclusion}
{\name} sets a new state-of-the-art for Rashi-script OCR, and outperforms frontier LLMs as well. We are pleased to release it to the NLP community. 


\end{document}